\documentclass[]{ceurart}
\usepackage{amsmath,amssymb,booktabs}
\usepackage{enumitem}
\setlist{nosep,leftmargin=*}
\begin{document}

\copyrightyear{2026}
\copyrightclause{Copyright for this paper by its authors. Use permitted under Creative Commons License Attribution 4.0 International (CC BY 4.0).}

\conference{MeMo Workshop on Mechanistic Interpretability \& Neuro-symbolic Approaches by-design, Rome (Italy), 24/6/2026}

\title{Towards Version-aware Operations and Transaction Memories for Multi-layer MeMo}

\author[1]{Peiran Li}[%
orcid=0009-0002-5807-3432,
email=peiran.li@fu-berlin.de,
]
\address[1]{Freie Universität Berlin,
Kaiserswerther Str. 16-18,
14195 Berlin,
Germany}

\begin{abstract}
MeMo proposes language models with explicit multi-layer correlation matrix memories (CMMs), where memorization, retrieval, and forgetting are architectural operations. This paper asks how such memories can reduce the need for retraining when knowledge changes. For changes expressible as MeMo memory associations, the model's accessible knowledge can be updated by editing explicit memories rather than retraining the whole model. We propose a version-aware operation layer in which high-level operations such as \textsc{replace}, \textsc{obsolete}, \textsc{keep-history}, \textsc{rollback}, and \textsc{trace} are compiled into MeMo-native primitive calls over sequences and tokens. The key observation is that a version-aware operation is rarely a single MeMo association. It is an ordered transaction of primitive edits, for example forgetting one sequence-token chain, memorizing another, preserving a historical chain, and recording an inverse program. The framework introduces two auxiliary CMMs: a Version CMM (V-CMM) for mapping version transitions to transaction handles, and a Transaction CMM (T-CMM) for storing reusable change contents and inverse programs. It supports both direct sequence-level edits and structured diff-level inputs, and outlines an evaluation route for update success, rollback, traceability, locality, and transaction reuse.
\end{abstract}


\maketitle

\section{Introduction}

Large Language Models are often adapted to new or changed knowledge through continued training, retrieval augmentation, or parameter-level model editing~\cite{gururangan2020dontstop,lewis2020rag,meng2022rome,meng2023memit}. These methods can be effective, but they are not ideal when a knowledge change is small and can already be expressed as an explicit memory association. In that case, the natural goal is to update the model's accessible knowledge by editing memory directly, rather than retraining on a fully revised corpus. MeMo is a useful starting point because it externalizes memory through correlation matrix memories and makes memorization and forgetting explicit operations rather than only side effects of training \cite{zanzotto2025memo}. Its final architecture is Multi-layer MeMo: a stack of CMMs composes input sequences and supports retrieval from explicit memory structures.

This paper keeps that Multi-layer MeMo architecture intact. We ask what operation layer is needed when explicit memories are long-lived, versioned, and repeatedly edited. MeMo provides primitive actions such as memorizing, retrieving, and forgetting a sequence-value association, but knowledge evolution also needs replacement, historical preservation, obsolescence, trace, and rollback.

The central idea is to treat such operations as transactions over MeMo primitive edits. A transaction is not a single association $(S,y)$; it is an ordered program that may remove $(S,y_{old})$, add $(S,y_{new})$, save historical memories, and record an inverse. We therefore design a two-stage auxiliary memory. A Version CMM (V-CMM) maps version transitions to transaction handles, and a Transaction CMM (T-CMM) maps transaction handles to executable edits. MeMo CMMs store what is remembered; V-CMM and T-CMM store which memory edits belong to a version change and how those edits should be executed.

The proposed algebra has two input levels. At the lower level, it supports direct sequence-token updates, such as replacing the memorized continuation of a MeMo sequence. At the higher level, structured knowledge changes, for example ontology moves or obsoletions, are compiled into sets of sequence-token transactions. Change languages such as KGCL provide a realistic interface for such structured changes \cite{hegde2025change}, while the contribution remains a version-aware operation algebra for Multi-layer MeMo.

\section{Multi-layer MeMo as Substrate}

A CMM stores associations by outer products, following classical correlation matrix memory and related distributed-memory ideas~\cite{kohonen1972correlation,anderson1972simple,plate1995holographic}. For example,
\begin{equation}
    C = \sum_i k_i v_i^{\top}.
\end{equation}
A query multiplies the memory matrix, and forgetting subtracts a corresponding outer product. Multi-layer MeMo extends this mechanism with several CMM layers that compose input sequences into higher-level sequence representations \cite{zanzotto2025memo}. We denote the original stack by:
\begin{equation}
    \mathcal{C}_{\mathrm{MeMo}} = \{C^{(1)}, C^{(2)}, \ldots, C^{(L)}, C^{(out)}\}.
\end{equation}

At the interface level, we abstract the existing implementation through three primitive procedures:
\begin{align}
    &\mathrm{memo}(S,y), &
    &\mathrm{forget}(S,y), &
    &\mathrm{retrieve}(S),
\end{align}
where $S$ is an input sequence and $y$ is the associated token or value. The internal multi-layer CMM machinery decides how the relevant layers are updated. The operation layer introduced below delegates to these primitives instead of directly modifying the layer design.

\section{Version-aware Operations}

\subsection{Direct sequence-level operations}

The lowest-level input to the proposed algebra is already MeMo-native: a sequence-token edit. At the operation level, a primitive edit can be represented as:
\begin{equation}
    e_j = (S_j,y_j,\lambda_j), \qquad \lambda_j\in\{+1,-1\},
\end{equation}
where $\lambda_j=+1$ denotes memorization and $\lambda_j=-1$ denotes forgetting. When an implementation exposes layer-specific updates, the executable form is $(\ell_j,S_j,y_j,\lambda_j)$; otherwise $\ell_j$ is left implicit and the standard multi-layer MeMo routine decides which internal CMMs to update. A high-level operation is an ordered transaction of such edits:
\begin{equation}
    \tau = \langle e_1; e_2; \ldots; e_m\rangle.
\end{equation}
The order matters when later edits depend on continuations created by earlier ones.
For example, a direct replacement of a MeMo continuation is:
\begin{equation}
\label{eq:directreplace}
\begin{aligned}
\textsc{replace}(S,y_{old},y_{new})=\langle&(S,y_{old},-1);\
&(S,y_{new},+1)\rangle.
\end{aligned}
\end{equation}
This operation changes the current view but does not erase the old value as a historical fact. Version preservation adds explicit versioned sequences:
\begin{equation}
\label{eq:history}
\begin{aligned}
\textsc{keep-History}=\langle&(S_{v_t},y_{old},+1);\
&(S_{v_{t+1}},y_{new},+1)\rangle.
\end{aligned}
\end{equation}
Thus, forgetting from the current view is not necessarily erasure. It can mean deactivation from the latest view while preserving earlier memories for version-specific queries and rollback.

\noindent\textbf{Multi-token continuations.}
Although examples are sometimes written as if $y$ were a single value, MeMo's native primitive is next-token memory. A continuation $Y=[y_1,\ldots,y_n]$ must therefore be decomposed into a chain of primitive edits:
\begin{equation}
\begin{aligned}
\mathrm{chain}(S,Y,\lambda)=\langle&(S,y_1,\lambda);(S y_1,y_2,\lambda);\\
&\ldots;(S y_1\cdots y_{n-1},y_n,\lambda)\rangle.
\end{aligned}
\end{equation}
A replacement between continuations becomes:
\begin{equation}
\textsc{replace}(S,Y_o,Y_n)=\mathrm{chain}(S,Y_o,-1);\mathrm{chain}(S,Y_n,+1).
\end{equation}
The T-CMM records either this expanded edit set for small transactions or, in the scalable implementation, a reusable template and parameter handles from which the edit chain is generated at execution time. In both cases, the original MeMo stack still receives only standard $(sequence,token)$ operations.

\subsection{Structured keys and higher-level changes}

A structured assertion can be represented as $(v,s,r,o)$, where $v$ is a version or view, $s$ is a subject, $r$ is a relation, and $o$ is an object value. To make it compatible with Multi-layer MeMo, the structured key is serialized into an input sequence:
\begin{equation}
\label{eq:sigma}
    \sigma(v,s,r) = [\texttt{VERSION},v,\texttt{SUBJECT},s,\texttt{RELATION},r].
\end{equation}
For structured resources, $s$, $r$, and $o$ denote stable identifiers rather than mutable labels; labels are used only for verbalization.
Writing the assertion is a standard MeMo call:
\begin{equation}
    \mathrm{memo}(\sigma(v,s,r),o),
\end{equation}
and removing it from a view is:
\begin{equation}
    \mathrm{forget}(\sigma(v,s,r),o).
\end{equation}
A structured change such as a class move or definition change is first normalized into one or more sequence-level transactions. For instance, replacing the latest value of $(s,r)$ from $o_{old}$ to $o_{new}$ becomes:
\begin{equation}
\label{eq:structuredreplace}
\begin{aligned}
\langle&(\sigma(\mathrm{v_{latest}},s,r),o_{old},-1);\
 &(\sigma(\mathrm{v_{latest}},s,r),o_{new},+1);\
 &(\sigma(v_t,s,r),o_{old},+1);\
 &(\sigma(v_{t+1},s,r),o_{new},+1)\rangle.
\end{aligned}
\end{equation}
The same mechanism supports \textsc{obsolete}: an old current status is deactivated, a new obsolete status is memorized, an optional replacement is memorized, and the previous status is preserved under the source version. 

\section{Version and Transaction Correlation Memories}

\subsection{Why CMMs instead of only logs?}

A version-aware operation is not itself a standard MeMo memory entry. MeMo stores next-token associations of the form $(S,y)$. A transaction may instead relate two continuations, such as replacing $Y_o$ by $Y_n$, and must be decomposed into several primitive $(sequence,token)$ edits. It may remove old edits, add new edits, preserve versioned edits, and store an inverse program.

An exact log is still useful, and in safety-critical settings it should remain the deterministic source of truth. However, logs alone do not provide a MeMo-native representation of reusable change contents. Version histories may be large, and many changes instantiate the same patterns: replacing a continuation, moving a class, obsoleting a term, or adding a synonym. We therefore separate version indexing from transaction content. A Version CMM (V-CMM) maps a version transition to transaction handles. A Transaction CMM (T-CMM) maps a transaction handle to the executable edit fields needed by MeMo. The exact log guarantees replayability; V-CMM and T-CMM provide compact associative indexing and reuse.

\subsection{V-CMM for version indexing}

Let $v_a\rightarrow v_b$ be a version transition containing transactions $\tau_{a,b,1},\ldots,\tau_{a,b,n}$. A slot-based V-CMM stores the association between a version transition, a slot, and a transaction handle:
\begin{equation}
\label{eq:vcmm}
    V = \sum_{(v_a,v_b,i,\tau)\in\mathcal{V}}
    A(v_a,v_b,i)B(\tau)^{\top}.
\end{equation}
Querying $V$ with $A(v_a,v_b,i)$ retrieves the $i$-th transaction handle for that transition. A count entry or a stop symbol can indicate how many slots are active. This slot-based form is easier to audit and execute deterministically than compressing all transactions into a single vector. A more compact variant could bundle all transaction vectors for a transition into a superposed representation, but we treat that as an associative retrieval optimization rather than the execution semantics.

\subsection{T-CMM for executable change content}

A transaction entry is represented as a mapping:
\begin{equation}
    (\tau,j,a) \mapsto b,
\end{equation}
where $\tau$ is a transaction identifier, $j$ indexes a primitive edit inside the transaction, $a$ is a field name, and $b$ is the field value. Examples include:
\begin{align}
    &(\tau,j,\mathrm{sequence}) \mapsto S_j, &
    &(\tau,j,\mathrm{value}) \mapsto y_j, \\
    &(\tau,j,\mathrm{sign}) \mapsto \lambda_j, &
    &(\tau,j,\mathrm{layer}) \mapsto \ell_j.
\end{align}
The T-CMM is:
\begin{equation}
\label{eq:tcmm}
    T = \sum_{(\tau,j,a,b)\in\mathcal{T}}
    \Gamma(\tau,j,a)\Omega_a(b)^{\top},
\end{equation}
where $\Gamma$ encodes transaction-field keys and $\Omega_a$ is a field-dependent value encoder. This field dependence is necessary because not all transaction values have the same role. Metadata fields such as operation type, source version, or template identifier may use ordinary typed symbolic encoders. Executable fields, however, must be MeMo-compatible: sequence fields are decoded into sequences or handles that can be passed to the relevant MeMo sequence encoder $k^{(\ell)}$, token fields are decoded into values compatible with the MeMo value encoder $v^{(\ell)}$, and control fields such as sign and layer use separate typed encoders.

Decoding a transaction therefore produces an ordered executable program:
\begin{equation}
\label{eq:decode}
    \mathrm{decode}_T(\tau)=\langle(\ell_j,S_j,y_j,\lambda_j)\rangle_{j=1}^{m}.
\end{equation}
Here $\ell_j$ may specify a target MeMo layer, or be omitted when the standard multi-layer MeMo memorization/forgetting routine should decide how to update its internal CMM stack. Execution delegates to MeMo primitives as an ordered sequence:
\begin{equation}
\label{eq:execute}
    \mathrm{execute}(\tau)=p_1; p_2; \cdots; p_m,
\end{equation}
where
\begin{equation}
\label{eq:primitive-call}
    p_j=
    \begin{cases}
    \mathrm{memo}(S_j,y_j), & \lambda_j=+1,\\
    \mathrm{forget}(S_j,y_j), & \lambda_j=-1.
    \end{cases}
\end{equation}
If an implementation exposes layer-specific primitives, $p_j$ can additionally carry $\ell_j$ and apply the corresponding MeMo-compatible encoders $k^{(\ell_j)}$ and $v^{(\ell_j)}$. Otherwise, the existing Multi-layer MeMo procedure performs the layer updates internally.

Rollback is obtained by reversing the ordered program and flipping signs:
\begin{equation}
\label{eq:inverse-transaction}
    \tau^{-1}=\langle(\ell_m,S_m,y_m,-\lambda_m);\ldots;(\ell_1,S_1,y_1,-\lambda_1)\rangle.
\end{equation}
Rollback has three cases. If $\tau$ is isolated or the last-applied transaction, the inverse program can be applied directly. If transactions are dependent, the system should replay the exact log to the target version. If versions are represented as logical views, rollback can instead switch the latest-view alias to an earlier version without physically subtracting memories.

\subsection{Templates and reuse}

Repeated changes should not require storing fully expanded edit programs each time. The T-CMM can store transaction templates and compact parameters. A minimal \textsc{replace} template is:
\begin{equation}
\label{eq:replace-template}
\mathrm{template}_{\mathrm{rep}}=[(\mathrm{slot}_S,\mathrm{slot}_{old},-1),(\mathrm{slot}_S,\mathrm{slot}_{new},+1)],
\end{equation}
and concrete edits are obtained by instantiation,
\begin{equation}
\label{eq:instantiate-template}
\mathrm{instantiate}(\mathrm{template}_{\mathrm{rep}},S,y_{old},y_{new}).
\end{equation}
Longer templates add source-version and target-version slots for history preservation. Thus, V-CMM indexes which transactions belong to a version transition, while T-CMM stores reusable change contents rather than duplicating full logs for every version.

\section{Examples}

\subsection{Direct sequence-level replacement}

Suppose a MeMo model has a current memory reflecting an older classification:
\begin{equation}
    S=\text{``Pluto is a''} \rightarrow [\mathrm{planet}].
\end{equation}
A later knowledge version classifies Pluto as a dwarf planet:
\begin{equation}
    S \rightarrow [\mathrm{dwarf},\mathrm{planet}].
\end{equation}
This example enters the algebra directly at the sequence-continuation level. For the transition $v_1\rightarrow v_2$, V-CMM first indexes the associated transaction handle:
\begin{equation}
    A(v_1,v_2,1)^\top V \approx B(\tau_{pluto}).
\end{equation}
T-CMM then decodes $\tau_{pluto}$ as an instance of a reusable continuation-replacement template with parameters $(S,[\mathrm{planet}],[\mathrm{dwarf},\mathrm{planet}],v_1,v_2)$. Because the new continuation contains two tokens, the transaction cannot be a single MeMo write. It is expanded into MeMo-compatible primitive edits:
\begin{align}
    &e_1=(\ell,S,\mathrm{planet},-1), \\[-1mm]
    &e_2=(\ell,S,\mathrm{dwarf},+1), \\[-1mm]
    &e_3=(\ell,S\ \mathrm{dwarf},\mathrm{planet},+1).
\end{align}
Historical preservation adds versioned chains:
\begin{align}
    &\mathrm{chain}(S_{v_1},[\mathrm{planet}],+1), \\[-1mm]
    &\mathrm{chain}(S_{v_2},[\mathrm{dwarf},\mathrm{planet}],+1).
\end{align}
Thus V-CMM routes the version transition to the transaction, T-CMM stores how the transaction expands, and the original MeMo stack only receives ordinary $(sequence,token)$ edits. The latest view generates the continuation $[\mathrm{dwarf},\mathrm{planet}]$, while a $v_1$ query generates $[\mathrm{planet}]$. For readability, continuations are shown as word-level tokens; an implementation would use the tokenizer units exposed to MeMo.

\subsection{Structured diff-level update}

A structured source can enter one level higher. Consider an ontology-style change:
\begin{equation}
    (v_1,\mathrm{Asthma},\mathrm{direct\_parent})\mapsto\mathrm{RespiratoryDisease}
\end{equation}
\begin{equation}
    (v_2,\mathrm{Asthma},\mathrm{direct\_parent})\mapsto\mathrm{ChronicRespiratoryDisease}
\end{equation}
A KGCL-like input may describe this as a class move. KGCL is a source-level change description: it records what changed in the ontology, but not which MeMo memories to edit. V-CMM is instead an execution-level index over MeMo-facing transactions. The compiler maps the source change to a transaction handle, for example:
\begin{equation}
    A(v_1,v_2,1)^{\top}V \approx B(\tau_{asthma}).
\end{equation}
The corresponding T-CMM entry stores a \textsc{replace-continuation} template with handles for the serialized sequence $\sigma(v,s,r)$, the old continuation, the new continuation, and the source/target versions. Decoding $\tau_{asthma}$ yields the same kind of ordered primitive edit program as in the Pluto example. Thus structured diffs are compiled into transactions indexed by V-CMM, decoded by T-CMM, and executed by the unchanged Multi-layer MeMo memorization and forgetting routines.

\section{Evaluation Roadmap and Scope}

The proposed work should be evaluated as a staged research programme. A first prototype will expose the native \(\mathrm{memo}\), \(\mathrm{forget}\), and \(\mathrm{retrieve}\) calls of an existing Multi-layer MeMo implementation, implement the operation algebra, and add V-CMM/T-CMM modules. It should cover direct continuation replacements and structured source-level changes such as ontology moves, synonym changes, obsoletions, and definition updates.

A small benchmark can use controlled toy facts for direct MeMo edits and ontology-style diffs, such as the Asthma parent-change example in Section~5. Baselines should include no update, retrieval augmentation, symbolic log-only patching without T-CMM reuse, continued training or retraining where feasible, and parameter editing methods for single-fact updates. The main metrics are update success, outdated-current suppression, historical preservation, rollback correctness, trace correctness, locality, and reuse. Trace is an interpretability contribution: it should expose the memory-level path from a current answer to the transaction, source version, old value, new value, and primitive edits that produced it.

\noindent\textbf{Scope and capacity.} The proposal targets local changes that can be expressed as MeMo-compatible memory edits; full rewrites may still require training. V-CMM indexes version deltas rather than full versions: for \(v_i\rightarrow v_{i+1}\), let \(\Delta_i=\langle\tau_{i,1};\ldots;\tau_{i,n_i}\rangle\) with \(n_i\leq B\). Large changes can be split into patch batches, and long transactions into bounded sub-transactions. T-CMM stores reusable templates and parameter handles rather than fully expanded edit programs, so a bounded delta adds \(O(BP)\) transaction-memory parameters, excluding symbol tables and the exact log.

\section{Conclusion}

This paper formulates a version-aware operation layer for Multi-layer MeMo. The original architecture remains unchanged: V-CMM indexes version transitions to transaction handles, T-CMM stores reusable change contents, and the MeMo stack executes the decoded memo/forget calls. The research agenda is to test whether explicit associative memories can support incremental, reversible, and traceable knowledge updates in cases where the relevant change can be expressed as MeMo-compatible memory associations. The same transaction mechanism may also support future authorship and provenance attribution by associating memory edits with authors, curators, documents, or source communities, extending traceability from what changed to who introduced a memory and which source justified it.

\section*{Declaration on Generative AI}

Generative AI tools were used as writing and editing assistance during the preparation of this paper, including support for language polishing, structural revision, and improving clarity and conciseness. The author takes full responsibility for the content of the paper.

\begingroup
\small
\bibliography{references}
\endgroup

\end{document}